\documentclass[letterpaper, 10 pt, conference]{ieeeconf}  

\IEEEoverridecommandlockouts                              

\usepackage{graphics} 
\usepackage{epsfig} 
\usepackage{mathptmx} 
\usepackage{times} 
\usepackage{amsmath} 
\usepackage{amssymb}  
\usepackage{hyperref}

\usepackage{cleveref}
\usepackage{graphicx}
\usepackage{subcaption}

\newcommand{\context}{context generator}

\newcommand{\llmplanner}{LLM planner}
\newcommand{\controller}{controller}
\newcommand{\Controller}{Controller}
\newcommand{\housekeep}{Housekeep}
\newcommand{\ourmethod}{LLM-Personalize}

\newcommand{\imitate}{self-training}
\newcommand{\gpt}{GPT-3.5-turbo}
\usepackage{color}
\usepackage{booktabs}
\usepackage{multirow} 
\usepackage[normalem]{ulem}
\useunder{\uline}{\ul}{}
\usepackage{tabularx}
\usepackage[utf8]{inputenc}
\usepackage{pifont}
\usepackage{newunicodechar}
\newunicodechar{✓}{\ding{51}}
\newunicodechar{✗}{\ding{55}}
\newcommand{\si}{ST}
\newcommand{\il}{IL}
\newcommand{\ione}{iter=1}
\newcommand{\itwo}{iter=2}
\usepackage{textcomp}

\newcommand{\obj}{\emph{obj}}
\newcommand{\rec}{\emph{rec}}
\newcommand{\room}{\emph{room}}

\title{\LARGE \bf
\ourmethod{}: Aligning LLM Planners with Human Preferences via Reinforced Self-Training for Housekeeping Robots}

\author{Dongge Han$^{1}$, Trevor McInroe, Adam Jelley, Stefano V. Albrecht, Peter Bell, Amos Storkey
\thanks{$^{1}$All authors are affiliated with the School of Informatics, University of Edinburgh, Edinburgh, UK. Correspond to {\tt\small dongge.han@ed.ac.uk.}}%
}

\begin{document}

\maketitle
\thispagestyle{empty}
\pagestyle{empty}

\begin{abstract}
Large language models (LLMs) have shown significant potential for robotics applications, particularly task planning, by harnessing their language comprehension and text generation capabilities. However, in applications such as household robotics, a critical gap remains in the personalization of these models to individual user preferences. We introduce \ourmethod{}, a novel framework with an optimization pipeline designed to personalize LLM planners for household robotics. Our \ourmethod{} framework features an LLM planner that performs iterative planning in multi-room, partially-observable household scenarios, making use of a scene graph constructed with local observations. The generated plan consists of a sequence of high-level actions which are subsequently executed by a controller.
Central to our approach is the optimization pipeline, which combines imitation learning and iterative self-training to personalize the LLM planner. In particular, the imitation learning phase performs initial LLM alignment from demonstrations, and bootstraps the model to facilitate effective iterative self-training, which further explores and aligns the model to user preferences. We evaluate \ourmethod{} on Housekeep, a challenging simulated real-world 3D benchmark for household rearrangements, and show that LLM-Personalize achieves more than a 30 percent increase in success rate over existing LLM planners, showcasing significantly improved alignment with human preferences. Project page: \url{https://donggehan.github.io/projectllmpersonalize/}.
\end{abstract}

\section{INTRODUCTION}
The application of large language models (LLMs) to the robotics domain has demonstrated substantial potential, especially in the realm of task planning~\cite{song2023llm, ahn2022can, rana2023sayplan, huang2022language, liang2023code, mai2023llm, huang2022inner, huang2023grounded}, by leveraging their advanced language comprehension and text generation capabilities. An important challenge of using LLM-powered planners is the alignment of the LLM with the specific task context. While many studies have focused on grounding LLM planners to the physical contexts of the tasks to ensure executability of the generated plans and their relevance to the environment, our work aims to further extend this foundation to study personalization, an important aspect to household robotics which tailors the functionality of the LLM planner to the unique household needs and preferences. 

\begin{figure}[t]
    \centering
    \includegraphics[width=1\linewidth]{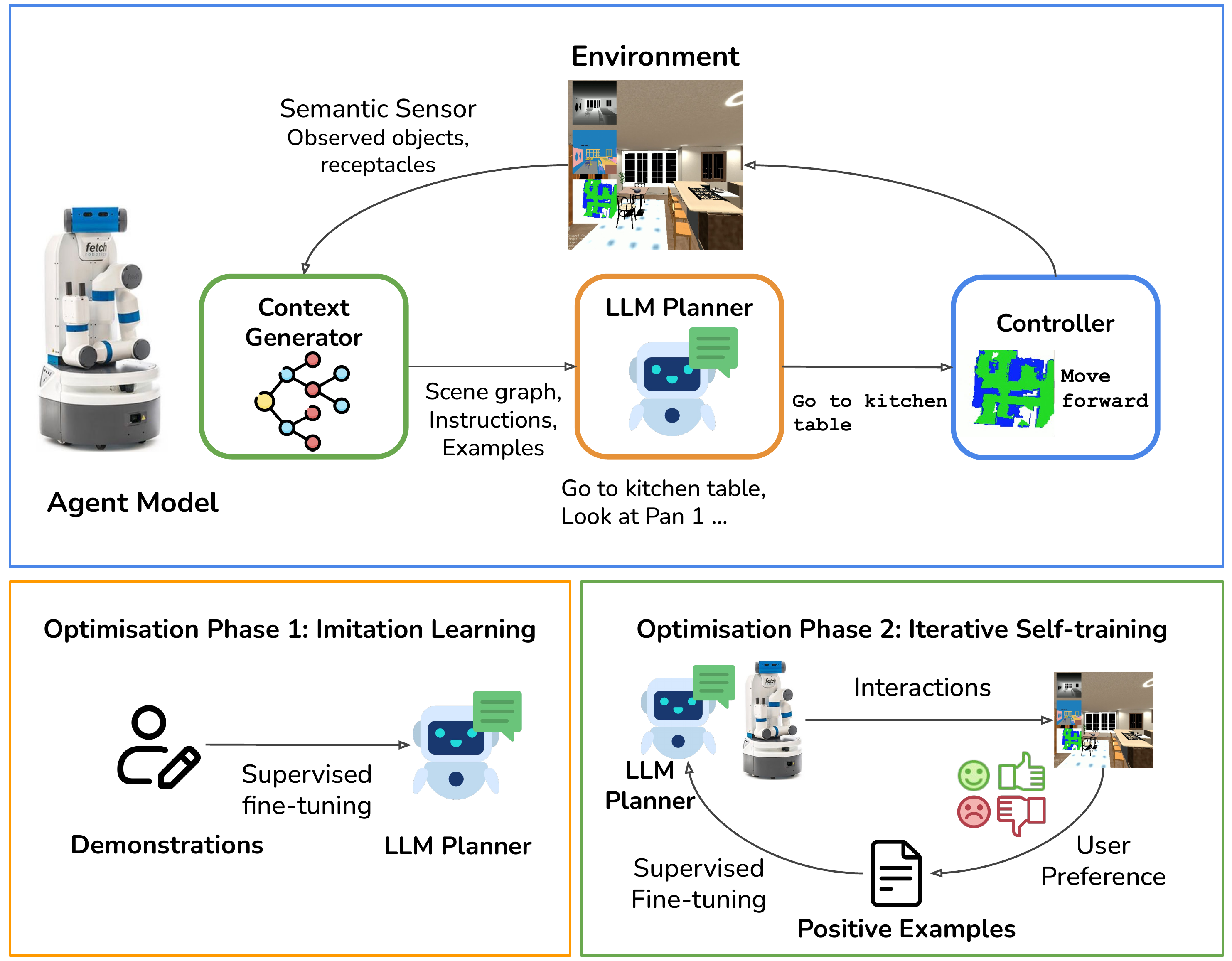}
    \caption{Illustration of \ourmethod{}. Agent architecture: The Context Generator constructs and updates a scene graph from local observations. The LLM Planner uses the graph to produce a plan as a sequence of high-level actions, and iteratively re-plans when the previous plan has been executed. Each high-level action is translated to a sequence of control actions and executed by the Controller. To personalize the LLM Planner, we introduce an optimization pipeline integrating imitation learning and iterative reinforced Self-Training to fine-tune and align the planner with user preferences.}
    \label{fig:model}
    \vspace{-7mm}
\end{figure}

Prior works on LLM grounding include aligning LLMs with the tasks' physical context through methods such as translating LLM generated plans to executable actions~\cite{huang2022language}, integrating context information such as affordance~\cite{ahn2022can, huang2023grounded}, scene graph~\cite{rana2023sayplan} or environment feedback~\cite{rana2023sayplan, huang2022inner}. Despite these advancements, there is noticeable gap in grounding LLM planners to personalized household preferences due to the inherent misalignment between the general-purpose LLM knowledge, designed to reflect common preferences, and the unique household preferences, e.g., one household may prefer a coffee mug to be placed on the dining table, whereas another may prefer for it to be in a kitchen cabinet.

To address this, we propose LLM-Personalize, a household robotic agent framework that performs object rearrangements in multi-room and partially observable household scenarios. As shown in Fig.~\ref{fig:model}, the model integrates three key components: context generator, LLM planner, and low-level controller. Central to personalizing the LLM planner to user preferences is our novel optimization pipeline that combines imitation learning with reinforced Self-Training (ReST)~\cite{gulcehre2023reinforced}. 
In the first phase, imitation learning is used to bootstrap the model in order to 1) guide the LLM planner to interpret complex input contexts, 2) initial alignment of the planner's behavior with example user preferences, 3) bootstrap the LLM planner to generate plans that can be straightforwardly annotated with user preferences, thus facilitating effective \imitate{} in the second phase, where the LLM planner further explores by collecting datasets of interactions, and refines itself based on the positive interactions according to the user preferences. 

We evaluate LLM-Personalize on Housekeep~\cite{kant2022housekeep}, a challenging, long-horizon, partially observable household rearrangements task suite, featuring diverse house layouts and a wide variety of receptacles and objects. The quality of object rearrangements are assessed according to the rearrangement success according to the Housekeep benchmark based on their collected human preference data. We demonstrate that LLM-Personalize outperforms state-of-the-art baseline LLM planners~\cite{song2023llm, ahn2022can, rana2023sayplan} with over a 30 percent increase in success rate, as a result of improved understanding and alignment with human preferences. 

\section{RELATED WORKS}
\textbf{LLM-Empowered Robotic Agents}
Recent works in task planning have effectively utilised pre-trained LLMs for generating executable plans for robotic agents~\cite{song2023llm, ahn2022can, rana2023sayplan, huang2022language,  liang2023code, mai2023llm, huang2022inner, huang2023grounded}. 
However, two key challenges that remain are the scalability of these methods to long-horizon planning tasks in large scenes, and misalignment of the LLM with the human preferences. \citet{wu2023tidybot} used LLMs for inferring rules summarizing personalized user preferences. In contrast, our work studies direct optimization and personalization of LLM planners for complex planning in multi-room household scenarios. Closely related to our work,~\citet{ahn2022can} performs grounding of LLM planners with affordance functions. However, it is applicable to small scenes and limited vocabulary of objects.~\citet{rana2023sayplan} addresses the scalability problem with a static scene graph.~\citet{song2023llm} addresses the scalability problem by allowing LLMs to plan iteratively. 
In this work, we directly use LLM for plan generation similarly to~\cite{song2023llm}. To address the long-horizon planning and large scene problem, our agent starts from an empty graph and dynamically updates the graph as it explores the house, and iteratively re-plan when the current plan finishes.\\ 
\textbf{Aligning LLMs with Human Preference}
Recent progress in LLM alignment are achieved via reinforcement learning (RL)~\cite{ouyang2022training, rafailov2023direct, glaese2022improving, akyurek2023rl4f}, or supervised learning (SL)~\cite{dong2023raft, xu2022leashing, liu2023chain, scheurer2023training}. Notably reinforced Self-Training (ReST)~\cite{gulcehre2023reinforced} can utilize either an RL or SL objective. 
In this work, we adopt (supervised) imitation learning (IL) to bootstrap our LLM planner. Then we adapt ReST to our LLM planner which iteratively explores and aligns itself with human preferences via supervised fine-tuning. 
While pairwise-comparison methods like Direct Preference Optimization (DPO)~\cite{rafailov2024direct} could theoretically be used for fine-tuning the LLM planner, they require paired positive and negative responses for each prompt, which is not supported by our human preference dataset or use case. Moreover, DPO demands custom loss functions and access to model gradients, not supported by online LLM fine-tuning APIs (e.g. GPT-3.5). In contrast, ReST circumvents these limitations. Therefore, we chose to use ReST for its simplicity and versatility, leaving exploration of other feedback approaches for future work.

\section{METHOD}
\begin{figure*}[t]
    \centering
    \includegraphics[width=0.95\textwidth, height=0.29\textwidth]{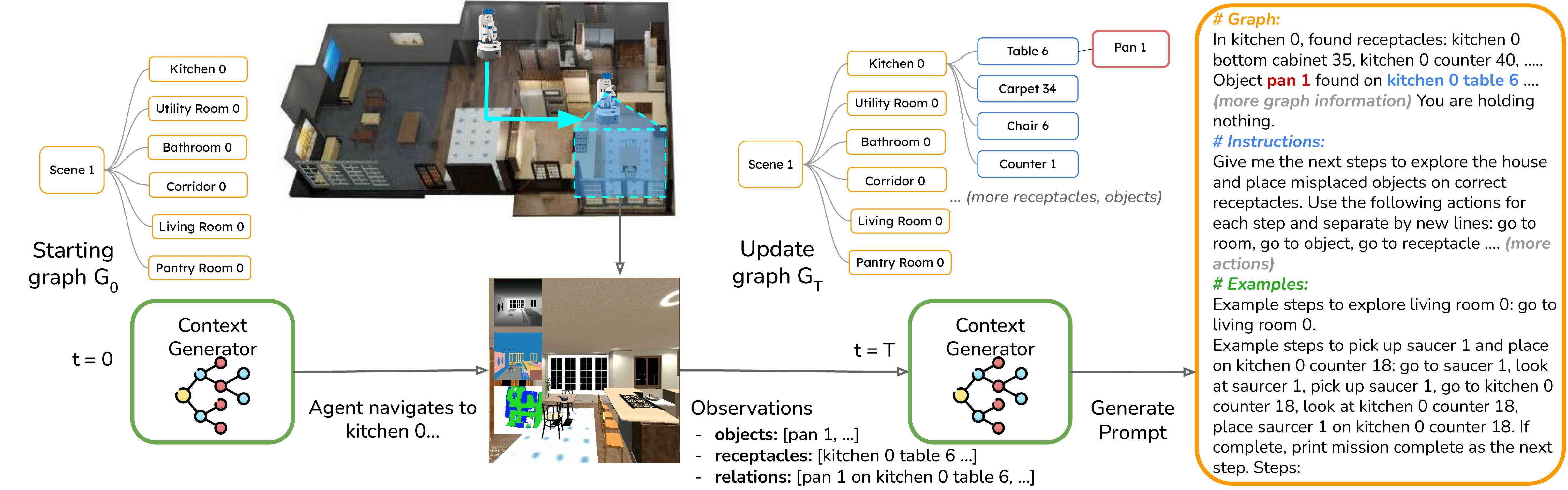}
    \caption{The Context generator builds and updates the graph of the household state of rooms, receptacles and objects, derived from the robot's local observations at each timestep. The information is provided as a prompt to the \llmplanner{}. Top-down view of the scene is for illustration only, the robot only has access to the 1st-person view.}
    \label{fig:context_generator}
    \vspace{-3mm}
\end{figure*}

\subsection{Problem Formulation} \label{sec:problem_formulation}
In this paper, we address the lack of personalization of LLM planners in household robotics tasks, for this purpose, we use Housekeep~\cite{kant2022housekeep}, a collection of 3D simulated household tasks, where a robot is tasked to rearrange misplaced objects to suit collected user preferences. A \emph{\textbf{scene}}, as shown in Fig.~\ref{fig:context_generator} and Fig.~\ref{fig:scenes}, refers to a household layout which includes a selection of \emph{\textbf{rooms}} and an arrangement of \emph{\textbf{receptacles}} in each room.  Let $j$ count through each receptacle, and $\rec_j$ denote high-level information about each receptacle (unique id, receptacle type, and the room it is in, e.g., \emph{kitchen 0 table 6}). 
Likewise, let $i$ count through each of the \emph{\textbf{objects}}, and let $\obj_i$ denote the high-level information about an object (its unique id, object type, e.g., \emph{laptop 1}).
Finally, we use $M_j^t$ to denote the set of indices $i$ of objects that are on a receptacle $j$ at time $t$ -- object locations will change over time. This set can be empty, and for some receptacles, $M^t_j$ may be constrained in cardinality. All the receptacles, objects and locations must be discovered by the agent. Their existence is not given a priori. Each \textbf{\emph{task/episode}} is of 1000 timesteps and includes a scene with a random selection of objects, some are misplaced on the wrong receptacles. The task for the agent is described as ``Give me the next steps to explore the house and place misplaced objects on correct receptacles''. The idea of misplacement has to be understood or learnt by the agent.

At each timestep $t$, the robot receives an egocentric (first-person) observation about a number of receptacles and any objects located therein. We collect the indices of the observed receptacles in $R_t$. So the observation at time $t$ consists of the high level observation $o_t=\{(\rec_j, \obj_i\mid i \in M^t_j)\mid j \in R_t\}$,
which is a list of the observed receptacles and the associated objects located on those receptacles, along with additional lower-level information, such as the coordinates of the objects etc. The robot can hold one object at a time and selects an action from its (low-level) action space $A = \{\, $\textit{move forward}, \textit{turn left}, \textit{turn right}, \textit{look up}, \textit{look down}, \textit{grab/release}$ \}$. It receives a reward $+1$ for placing an object on a correct receptacle, $-1$ for grabbing/removing an already correctly placed object, else $0$. The correctness of \obj{}-\rec{} placement is decided by a human preference dataset collected by \housekeep{} from human annotators. 

\subsection{Model} 
Our robotic agent model is designed to perform long-horizon planning in the partially observable household scenarios with three key components: the \context{}, the \llmplanner{} and the \controller{}. Specifically, the \context{} provides context information for decision-making in the form of prompts, by maintaining a graph of the household state derived from observations. For decision making, we choose a two-level design that integrates a high-level LLM planner and a \controller{} which executes the generated plans using low-level control actions. Given our primary focus on personalizing LLM planners, we use an off-the-shelf \controller{} from the simulator, and focus on designing the \context{} and \llmplanner{}.

\subsubsection{Context Generator}
The \context{} provides information of the current household context and useful instructions as a prompt for the downstream \llmplanner{}. Specifically, our \context{} provides three pieces of information: the current household state, instructions, and examples for in-context learning~\cite{brown2020language}. As the agent never observes the full state of the household, but only receives a partial view $o_t$, it is important that the agent maintains and refines an internal representation of the household state to correctly choose the placement of objects (e.g., only using local observations can lead to a suboptimal placement when the correct receptacle of a misplaced object is in a different room). To this end, our \context{} maintains a graph $G$ as shown in Fig.~\ref{fig:context_generator}. When a task starts, the \context{} initializes the graph $G_0$ of the house with empty room nodes. At each timestep $t$, the graph is updated with the locally observed objects and receptacles $o_t$ as the agent navigates around the house. To provide the information for decision-making, a prompt is constructed with a natural language description of $G_t$, the object held by the robot, instructions (a description of the overall rearrangement task, the role assigned to the LLM, and the available high-level actions), and two examples, one for room exploration and another one for moving an object from one receptacle to another, allowing the \llmplanner{} to follow via in-context learning.

\begin{figure*}[t]
    \centering
    \includegraphics[width=0.9\textwidth]{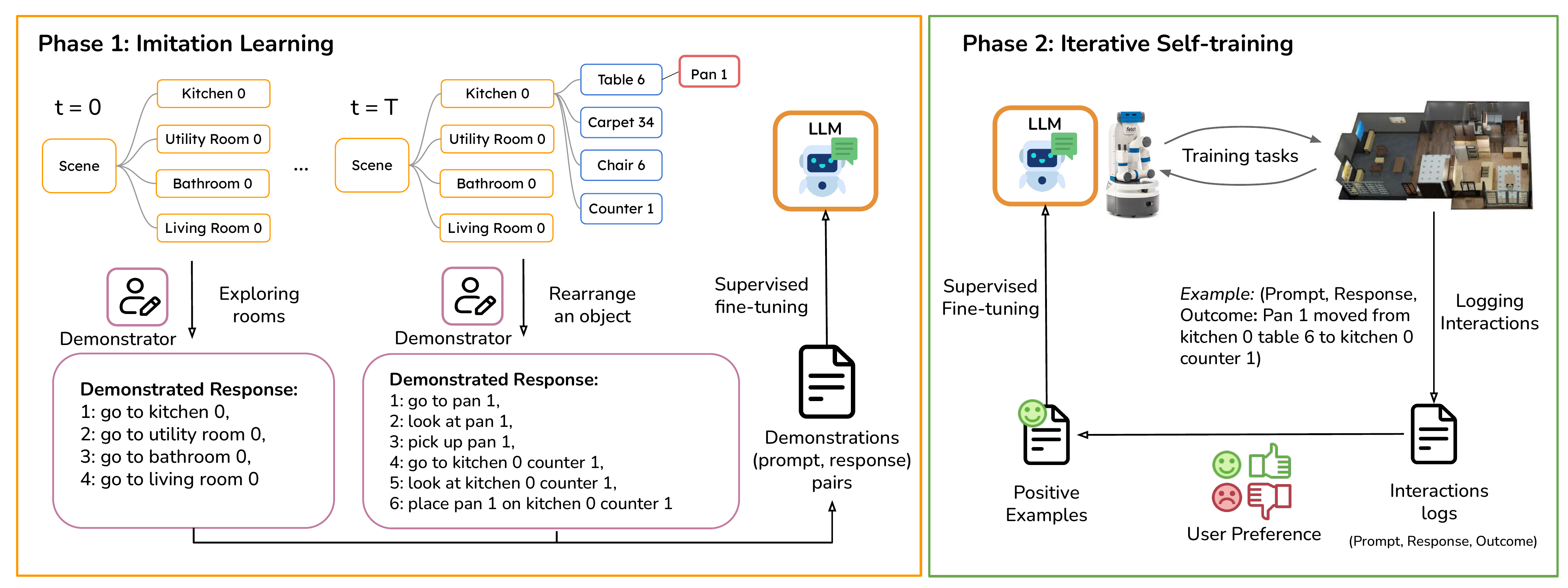}
    \caption{Optimization pipeline of \ourmethod{} using imitation learning and iterative reinforced~\imitate{}.}
    \label{fig:optimization}
    \vspace{-5mm}
\end{figure*}

\subsubsection{LLM Planner}
The \llmplanner{} is the core decision-making module that generates a \emph{high-level plan as a sequence of high-level actions}, and each high-level action is translated by the low-level \controller{} to low-level actions and executed. 
The set of available \emph{\textbf{high-level actions}} are: $\Omega = $
$\{$go to \obj{}/\rec{}/\room{},
look at \obj{}/\rec{},
pick up \obj{},
place \obj{} on \rec{}$\}$, where the \obj{}, \rec{}, \room{} are replaced by the actual names of these target entities, as introduced in~\cref{sec:problem_formulation}.
To handle partial observability in the multi-room households, we adopt an iterative planning procedure which enables the \llmplanner{} to re-plan when the previous plan has finished execution. This allows the agent to obtain a more comprehensive understanding of the household state while exploring and navigating the house, leading to improved rearrangement decisions. On the other hand, compared with single-step iterative planning where each plan includes only one high-level action, our approach builds cohesive plans that better account for the inter-dependencies and the cumulative effect of the action sequences. The iterative procedure works as follows: Denote $n \in \mathbb{N}$ as a high-level planning iteration. At iteration $n$, the \llmplanner{} receives a prompt from the \context{} and generates an immediate action \emph{plan} as a sequence of high-level actions $\omega\in\Omega$: $\textit{p}_{n} = (\omega_0, \omega_1, \ldots)$ (see Fig.~\ref{fig:optimization} for two example plans). The plan is sent to the \controller{} where the high-level actions are executed sequentially. Each high-level action is translated to a sequence of low-level control actions $a \in A$, e.g., \emph{go to pan 1 $\rightarrow$ (move forward, turn left, $\ldots$)}. Once the \controller{} finished executing all high-level actions in $\textit{p}_{n}$, say at timestep $t=T$, the next plan iteration $n+1$ starts where the \llmplanner{} is prompted again to generate a new plan $p_{n+1}$ and executed by the \controller{}. This process is repeated iteratively. 

The \llmplanner{} is implemented with an LLM model and a parser. Upon receiving a prompt from the \context{}, the LLM returns a plan in natural language as a sequence of high-level actions: \emph{go to pan 1, pick up pan 1, ...}. Specifically, the first 10 high-level actions from the sequence are used, as we often observe a decrease in quality towards the end of a long response from the LLM. The parser then extracts from each high-level action the target action (i.e., one of \emph{go to, look at, pick up, place}) and the target entities (i.e., \obj{}, \rec{}, \room{}) and send to the \controller{} to be translated and executed as low-level control actions.

\subsubsection{\Controller{}}
Given our primary focus on personalizing the LLM planner, we use the off-the-shelf \controller{} from the \housekeep{} simulator, which maps each high-level action to a sequence of low-level actions. More details can be found in the Appendix~\ref{subsec:controller_appendix}.

\subsection{Personalizing the LLM Planner}
\label{subsec:optimisation}

Despite our model architecture being well-suited to the partially observable household scenarios, we observed two challenges that necessitated a tailored optimization process. 
1) LLM planners struggle with effectively extracting precise information from complex input contexts (e.g., resulting in plans with partial object names). This is compounded by the complexity of accurately sequencing high-level actions to ensure executability. 
2) Misalignment between the LLM planners' decisions and the personalized preferences of users. 

Reinforced~\imitate{}\footnote{For simplicity, we'll refer to reinforced~\imitate{} (ReST) as~\imitate{} (\si{}) in the sections that follow.} provides a promising approach to optimizing and personalizing the \llmplanner{} with user preferences, by iteratively performing a \emph{grow} step where a training dataset is collected by prompting the LLM to generate multiple responses for each prompt, and an \emph{improve} step, where the dataset is filtered according to human preferences, followed by fine-tuning the LLM on the filtered dataset. However, direct application of \si{} to the \llmplanner{} presents new, unique challenges: Unlike single-step generation tasks, e.g., \si's initial application domain of machine translation), the household robotics tasks often involve long-horizon planning, where the \llmplanner{} may generate a plan consisting of both correct and incorrect placements actions, making it difficult to annotate with human preferences and extract clean training examples for automatic \imitate{}. 

To this end, we introduce a tailored optimization pipeline that integrates imitation learning and \imitate{}. The imitation learning phase bootstraps the \llmplanner{} to effectively interpret the complex context, produce executable plans, and perform initial alignment with example user preferences. Moreover, the demonstrations are designed to bootstrap the \llmplanner{} to generate plans that can be clearly annotated, thus facilitating effective \imitate{} in the second phase, which allows the \llmplanner{} to further explore and refine its planning strategies based on user preferences. 

\subsubsection{Imitation Learning}
As shown in Fig.~\ref{fig:optimization}, we build a demonstrator module to generate demonstrated responses for the \llmplanner{} on a set of demonstration tasks. On receiving a  prompt $\mathbf{x}$ from the \context{}, the demonstrator produces a plan $\mathbf{y}$ which either explores one of the rooms or rearrange a single object, using the scene graph from the \context{} and the correct object-receptacle mapping according to the human preference dataset (described in~\cref{sec:problem_formulation}). Specifically, when prompted at the start of a task, the demonstrator produces a plan of high-level actions to visit each of the rooms. After the plan is executed, the agent will have discovered some misplaced objects and receptacles in each room. The demonstrator will be prompted again to generate a plan, which rearranges one of the discovered misplaced object (picked randomly) to a discovered correct receptacle. The plan is executed by the \controller{}, and we iterate the procedure until all discovered objects are correctly placed. 

To bootstrap the \llmplanner{}, we prepare the collected demonstrations as pairs of prompts and target responses, $\mathcal{D}_\textnormal{demo}=\{$($\mathbf{x}^i$, $\mathbf{y}^i$)$\}_{i=1}^N$, where both $\mathbf{x}^i, \mathbf{y}^i$ are sequences of tokens,  the response $\mathbf{y}^i$ being a plan by the demonstrator. 
Given a pre-trained autoregressive LLM $P_\theta(\mathbf{y} \mid \mathbf{x})$ parametrized by $\theta$, we perform supervised fine-tuning on $\mathcal{D}_\textnormal{demo}$ by minimizing the negative log likelihood (NLL) loss:
\begin{equation}\label{eq:nll}
\resizebox{.85\hsize}{!}{$\mathcal{L}_{NLL} = -\mathbb{E}_{(\mathbf{x},\mathbf{y})\in \mathcal{D_\textnormal{demo}}} \left[\sum_{\tau=1}^{|\mathbf{y}|} \log P_{\theta}(\mathbf{y}_\tau \mid \mathbf{y}_{1:\tau-1}, \mathbf{x})\right]$}
\end{equation}

The above demonstration design guides the \llmplanner{} to more accurately extract information from complex input contexts and improve plan executability. More importantly, it ensures that each plan has a uniform objective (i.e, perform exploration or rearrange a single object), allowing the plan to be straightforwardly annotated with user preferences during \imitate{}. 

\subsubsection{Iterative Reinforced \imitate{}}
Next, we fine-tune the bootstrapped \llmplanner{} to further improve personalization via iterative \imitate{} on a set of training tasks. This allows the \llmplanner{} to explore more rearrangement options and improve its placement decisions through imitating the positive examples.
As shown in Fig.~\ref{fig:optimization}, to start a \imitate{} iteration, we use the \llmplanner{} to explore by collecting episodes of experiences on the training tasks and we log $M$ interactions as tuples of prompt, response and outcomes: $\{(\mathbf{x}^i, \mathbf{y}^i, out^i)\}_{i=1}^M$ (Example outcome: pan 1 moved from kitchen 0 table 6 to kitchen 0 counter 1). We annotate each prompt and response pair $(\mathbf{x}^i, \mathbf{y}^i)$ with the reward $r^i\in\{1,0,-1\}$ for the rearrangement outcomes according to the user preferences. 
Then, we collect a \imitate{} dataset by picking the prompt and response pairs with positive rewards $\mathcal{D}_\textnormal{self-train} = \{(\mathbf{x}^i, \mathbf{y}^i) \mid r^i > 0\}_{i=1}^M$
The final step of this \imitate{} iteration is to perform supervised fine-tuning of the bootstrapped LLM model over $\mathcal{D}_\textnormal{self-train}$ with the NLL objective as defined in Equation~\eqref{eq:nll}. This procedure is repeated iteratively where each \imitate{} iteration performs interaction collection and fine-tuning over the LLM obtained from the previous iteration.

\section{EXPERIMENTS}

\begin{figure*}[t]
    \centering
    \hfill
    \begin{subfigure}[b]{0.23\textwidth}
        \includegraphics[width=\textwidth, height=0.6\textwidth]{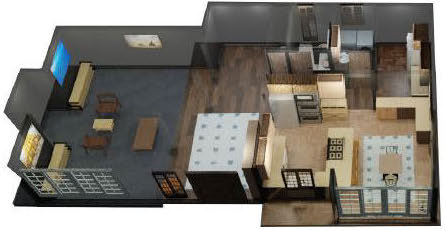}
        \caption{Scene 1 (kitchen, living room, corridor, bathroom, utility room, pantry room)}
        \label{fig:scene1}
    \end{subfigure}
    \hfill 
    \begin{subfigure}[b]{0.23\textwidth}
        \includegraphics[width=\textwidth, height=0.6\textwidth]{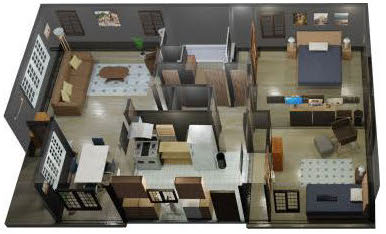}
        \caption{Scene 2 (kitchen, living room, dining room, child's room, bathroom, bedroom)}
        \label{fig:scene2}
    \end{subfigure}
    \hfill
    \begin{subfigure}[b]{0.22\textwidth}
        \includegraphics[width=\textwidth, height=0.7\textwidth]{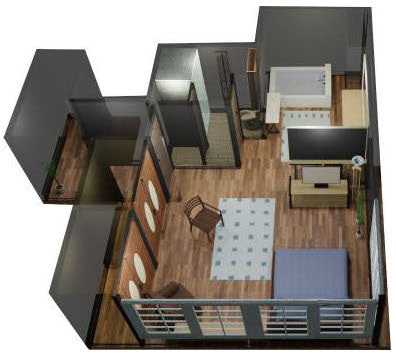}
        \caption{Scene 3 (corridor, bathroom, bedroom)}
        \label{fig:scene3}
    \end{subfigure}
    \hfill
    \begin{subfigure}[b]{0.21\textwidth}
        \includegraphics[width=\textwidth, height=0.7\textwidth]{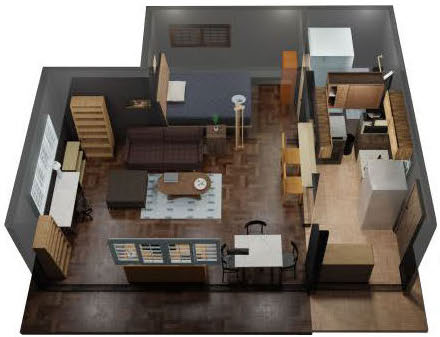}
        \caption{Scene 4 (kitchen, living room, bathroom, bedroom, lobby)}
        \label{fig:scene4}
    \end{subfigure}
    \hfill
    \vspace{-3mm}
    \caption{The Housekeep scenes used in our experiment.}
    \label{fig:scenes}
\end{figure*}

\begin{table*}[t]
\vspace{-3mm}
\resizebox{\linewidth}{!}{
    \centering
    \footnotesize
    \setlength\tabcolsep{1.6pt}
    \begin{tabular}{@{}cccccccccc@{}}
\toprule
\textbf{} & \textbf{Scene ID} & \multicolumn{2}{c}{\textbf{Scene 1}} & \multicolumn{2}{c}{\textbf{Scene 2}} & \multicolumn{2}{c}{\textbf{Scene 3}} & \multicolumn{2}{c}{\textbf{Scene 4}} \\ \midrule
 & \textbf{Task Set} & \textit{\textbf{train}} & \textit{\textbf{test}} & \textit{\textbf{train}} & \textit{\textbf{test}} & \textit{\textbf{train}} & \textit{\textbf{test}} & \textit{\textbf{train}} & \textit{\textbf{test}} \\ \midrule
\multirow{3}{*}{\textbf{Baselines}} & \textbf{SayCan} & -2.6 ± 1.9 & 0.0 ± 0.0 & -1.2 ± 1.2 & -10.6 ± 6.8 & -3.3 ± 2.2 & -8.0 ± 4.9 & -1.6 ± 1.6 & 0.0 ± 0.0 \\ \cmidrule(l){2-10} 
 & \textbf{SayPlan} & -7.0 ± 5.9 & -5.0 ± 5.0 & -6.8 ± 2.7 & -1.6 ± 9.2 & 0.4 ± 4.5 & -13.0 ± 8.3 & -10.7 ± 3.8 & -12.3 ± 15.3 \\ \cmidrule(l){2-10} 
 & \textbf{LLM-Planner} & 5.3 ± 4.4 & -3.6 ± 4.8 & -8.4 ± 3.8 & -9.8 ± 6.3 & -14.2 ± 4.0 & -4.0 ± 3.3 & -29.6 ± 5.2 & -30.2 ± 4.8 \\ \midrule
\multirow{3}{*}{\textbf{Ours}} & \textbf{LLM-Personalize (\il{})} & 4.1 ± 2.6 & 17.6 ± 6.1 & -3.3 ± 3.0 & 12.6 ± 9.0 & 22.6 ± 2.6 & 24.3 ± 4.4 & 10.9 ± 3.3 & \textbf{25.7 ± 6.7} \\ \cmidrule(l){2-10} 
 & \textbf{LLM-Personalize (\si{} \ione{})} & 17.9 ± 3.7 & 25.8 ± 6.6 & \textbf{19.4 ± 2.9} & 21.7 ± 5.6 & 32.4 ± 3.2 & 41.4 ± 6.3 & 24.2 ± 3.2 & 10.2 ± 6.7 \\ \cmidrule(l){2-10} 
 & \textbf{LLM-Personalize (\si{} \itwo{})} & \textbf{25.5 ± 3.1} & \textbf{29.6 ± 5.4} & 18.5 ± 2.6 & \textbf{25.2 ± 4.0} & \textbf{33.5 ± 3.8} & \textbf{43.3 ± 4.3} & \textbf{29.1 ± 2.9} & 20.4 ± 6.8 \\ \bottomrule
\end{tabular}
}
\caption{Average success rate on train and test sets across scenes. Each entry denotes the mean ± standard error of the mean across episodes. (\textbf{Boldface}: best variant across the task set, \textbf{\si{}}: self-training, \textbf{\il{}}: imitation learning)}
\label{table:main_results}
\vspace{-3mm}
\end{table*}
We aim to evaluate the hypothesis:
\emph{Optimizing \llmplanner{} through imitation learning and \imitate{} allows the \llmplanner{} to improve planning performance and alignment with user preferences.}
We demonstrate this through improved rearrangement success rate compared with baseline LLM planners and provide qualitative results that showcase the plans generated. 
Our ablations studies further evaluate plan executability, exploration and cross-domain (scene) generalisation of \ourmethod{} in the different training phases. 


\subsection{Experiment Setup}

We evaluate \ourmethod{} on 4 different scenes in \housekeep{}, each featuring a unique layout of rooms and receptacles, as shown in Fig.~\ref{fig:scenes}.
A \emph{task} is instantiated with a random selection of $5-10$ objects placed on different receptacles, among them $\sim 3-7$ objects are misplaced and needs to be rearranged. For each object, there is a list of correct receptacles not known to the agent. The task is challenging as the agent must explore the house, identify  misplaced objects and their correct receptacles, and avoid removing correctly placed objects.\\
\textbf{Evaluation metrics:}
We evaluate the agent performance via success rate~\cite{kant2022housekeep}, defined as the percentage of misplaced objects that are correctly re-arranged at the end of the task, among all misplaced objects at the start of the task. 
\begin{equation*}
\resizebox{.97\hsize}{!}{$\textit{\scriptsize Success Rate} = \left(\frac{\#\textit{correct at the end} - \#\textit{correct at the start}} {\#\textit{total misplaced objects at the start}}\right)$}
\end{equation*}
With the success rate defined in terms of difference, an agent is judged fairly for its correct and wrong placements: an agent which performed poorly that resulted in more misplaced objects at the end of the task will yield a negative success rate.
We measure the success rate on train and test tasks at each phase of the optimization pipeline. For each scene, we randomly sample three disjoint set of $10$ demonstration tasks, $20$ training tasks and $5$ test tasks. Hence the agent will encounter a random selection of objects and placement configurations on each task. During both training and testing, we collect experiences of $5$ episodes per task, and present the mean and standard error of the metrics across all collected episodes in the task sets. \\
\textbf{Architecture and Baselines}
We compare \ourmethod{} with the baseline LLM planning methods: LLM-Planner~\cite{song2023llm}, SayPlan~\cite{rana2023sayplan} and SayCan~\cite{ahn2022can}. For all methods, we use \gpt{} with temperature set to $1$ to reduce deterministic repetition in LLM responses (default range is $0 - 2$). The prompt includes the instruction, 
 graph description and two examples. We allow all methods to plan iteratively after the robot executed the previous plan. LLM-Planner with these configurations is adopted as the base model of \ourmethod{}, which we then optimize using imitation learning and \imitate{}. The fine-tuning as defined in Eq.\eqref{eq:nll} is performed via the OpenAI fine-tune API. For more details please refer to Appendix~\ref{subsec:baselines_appendix}.

\begin{figure*}
    \centering
    \includegraphics[width=0.95\textwidth, height=0.25\textwidth]{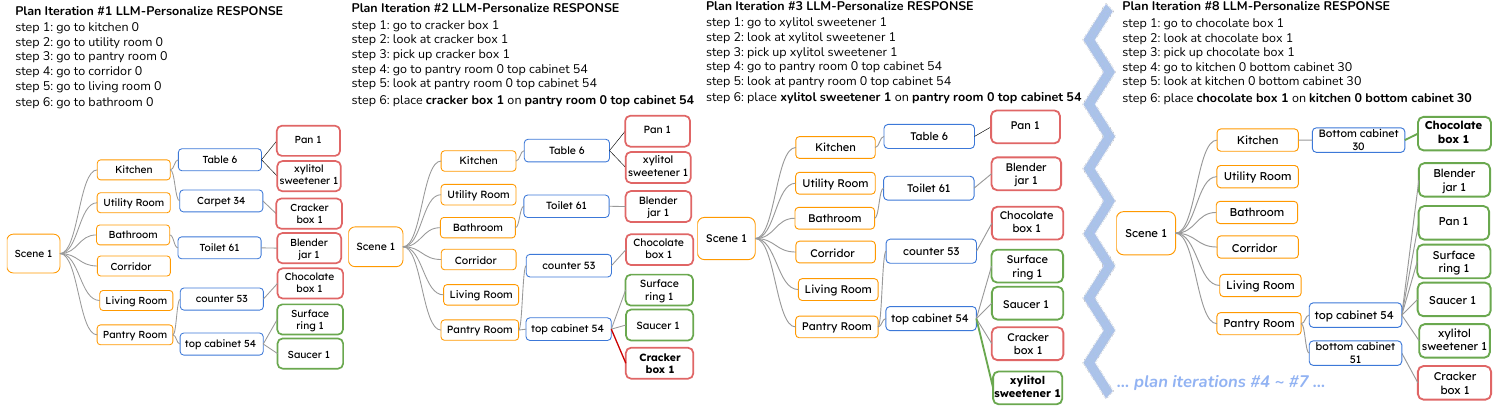}
    \vspace{-3mm}
    \caption{Demonstration of four planning iterations generated and executed by \ourmethod{} (top row) and the resulting graphs (bottom row) on a test task in \housekeep{}. Green/red object (leaf) nodes indicate correct/wrong placements. The object being moved is shown in boldface. This episode starts with 2 correctly placed objects and 5 misplaced objects (left), and changed to 6 correctly placed objects and only 1 misplaced objects after rearrangements (right). For clarity, the graphs only show receptacles with objects and omit all other receptacles.}
    \label{fig:qualitative_graph}
    \vspace{-3mm}
\end{figure*}

\subsection{Main Results}
\textbf{Quantitative Results}
We compare the average success rate of \ourmethod{} with the baselines in Table \ref{table:main_results}, and we show the effectiveness of our optimization framework by comparing the performance of \ourmethod{} at different optimization phases, namely, the base version (using LLM-Planner), imitation learning (\il{}), and various iterations of \imitate{} (\si{}). Each table entry shows the mean and standard error of the mean across all tasks in the train/test set and 5 runs per task. 
Overall, we observe that \ourmethod{} significantly outperforms all baseline methods across the tested scenes. For example, on the test set of Scene 1, the success rate of the baselines are near zero or negative, while \ourmethod{} achieved $29.6\%$ after imitation learning and two \imitate{} iterations. Similar trends can be observed across all scenes. After examining some detailed prompts and responses we identified that SayCan often has difficulty picking the best high-level action from a large number of available actions due to the large number of objects, receptacles and rooms. LLM-Planner is often able to produce correct pick and place action sequences following the in-context examples. Compared with LLM-Planner, SayPlan improves slightly (on 5 out of 8 task sets) as a result of improvement in plan executability due to revision with feedback. However, all baselines have difficulty knowing whether an object is misplaced or correctly placed, as well as the correct receptacles to place objects, due to lack of personalization. As a result, the negative scores across the baseline methods are often due to picking and placing correctly placed objects onto wrong receptacles. 

Comparing the different stages of \ourmethod{} on the test sets, we observe a general trend where the combination of imitation learning and \imitate{} lead to better results. First, bootstrapping from demonstrations improves over LLM-Planner (i.e., the base \ourmethod{} model). For example, on Scene 1, the success rate improved from $-3.6\%$ to $17.6\%$. This improvement is a result of improved executability due to action sequencing, better context understanding (e.g., agent correctly extracts and uses object names from the prompt), and initial alignment to personalized preferences shown in the demonstrations. Second, the \imitate{} iterations further improves performance with improved alignment with personalized preferences. For example, on Scene 1, the success rate after two iteration of \imitate{} grows from $17.6\%$ to $29.6\%$ compared with the imitation bootstrapped variant. We also observed that after bootstrapping, the model learns to explore object placements with improved accuracy during \imitate{} and with more \imitate{} iterations, the model exploits and commits to the learned correct placements while avoiding wrong ones.

Comparing the train and test performances of \ourmethod{}, we observe that the testing performance generally increases with increased training performance, except for Scene 3, where the test performance drops as a result of overfitting. This shows that our model is able to learn personalized preferences seen during training, and generalize to unseen object combinations and placements. \\
\textbf{Qualitative Results} 
In Fig.~\ref{fig:qualitative_graph}, we present plans generated by \ourmethod{} and executed by the robot on a test task in  \housekeep{} and the resulting graph of the household scene after each plan iteration. We can see that the agent learned to start by exploring the house, then rearrange one misplaced object at a plan iteration, and successfully rearranged 4 out of 5 misplaced objects.
\subsection{Ablation Studies}
In this section, we present ablation studies on \ourmethod{}'s plan executability, exploration vs. exploitation and cross-domain transfer performance.\\
\begin{figure}[tbp]

    \centering
    \begin{subfigure}[b]{0.49\columnwidth}
        \centering
        \includegraphics[width=0.95 
        \textwidth]{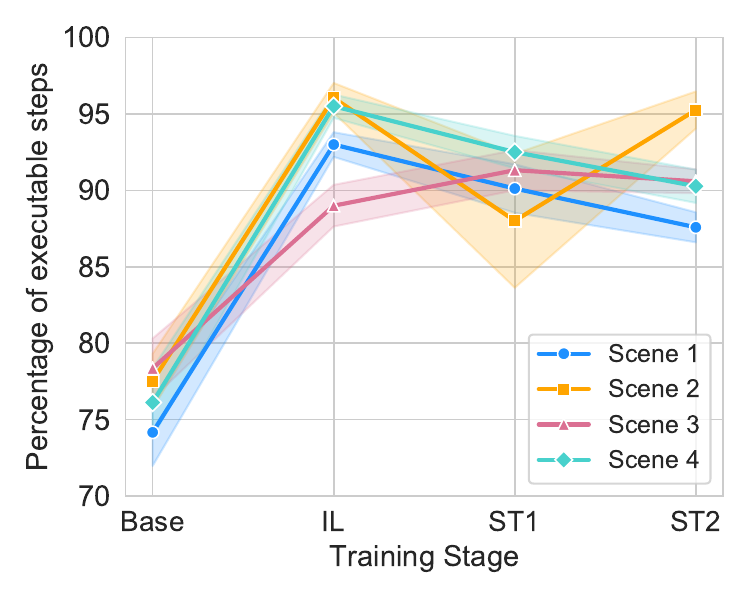} 
        \caption{Executability}
        \label{fig:executability}
    \end{subfigure}
    \hfill 
    \begin{subfigure}[b]{0.49\columnwidth}
        \centering
        \includegraphics[width=0.95\textwidth]{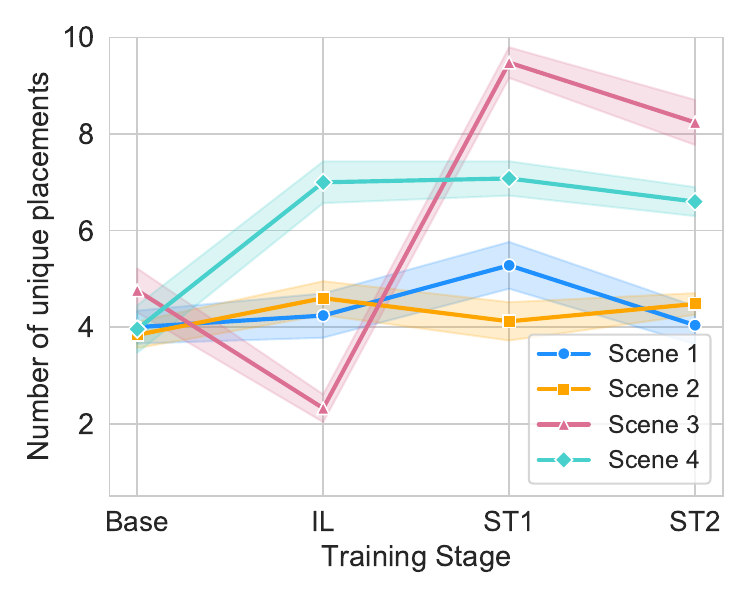} 
        \caption{Unique Placements}
        \label{fig:unique_placements}
    \end{subfigure}
    \caption{Ablations (a) percentage of executable high-level actions (b) unique placements executed. x-axis refers to \ourmethod{} at different phases -- Base: before optimization, IL: imitation learning, ST: Self-Training. Each point is an average value over 25 episodes (5 runs per task over 5 tasks in the test set) and shaded area refers to standard error of the mean.}
    \label{fig:ablations_1_2} 
    \vspace{-8mm}
\end{figure}
\textbf{Cross-domain Transfer Results}
In addition to the improved in-domain adaptation results in Table~\ref{table:main_results}, we show in Table~\ref{table:cross_domain} cross-domain transfer performance of \ourmethod{}. In this experiment, we train the model on a source scene (e.g., scene 2), and observe the performance change on the test set on a different scene (e.g., scene 1) with different rooms and receptacles. From the table, we can observe that through imitation learning and \imitate{}, \ourmethod{} is able to transfer to a different scene with improved test performance.\\
\textbf{Executability}
In Fig.~\ref{fig:executability} we present the executability improvement of \ourmethod{} (IL, \si1, \si2) compared to the base LLM-Planner (Base). Each point refers to the average percentage of high-level actions generated by the \llmplanner{} that are successfully executed per task. The \il{} bootstrapping significantly improved the planner's executability, due to improved context understanding and action sequencing, enabling \ourmethod{} to produce high quality training examples for \imitate{}, and we can observe consistently high executability from the \ourmethod{}(ST1) and \ourmethod{}(ST2) variants. \\
\textbf{Exploration vs. Exploitation} 
To analyse how the agent's exploration vs. exploitation behavior changes at different optimization phases, we show in Fig.~\ref{fig:unique_placements} the average number of unique placements executed per episode, where each unique placement refers to a pair of object and receptacle where the agent placed the object on the receptacle. 
Higher degree of exploration behavior is indicated by higher number of unique placements, and in contrary, lower number of unique placements indicates more exploitation behavior. Fig.~\ref{fig:unique_placements} shows an increase in exploration from LLM-Planner (Base) to \ourmethod{}(\il{}), partly due to improved plan executability. From \il{} to \si1, the agent further increased exploration, where it explores placements beyond the behaviors learned from demonstrations. For example, on Scene 3 the average unique placements increased from 2.32 to 9.48. Compared with \si1, \si2 typically shows more exploitation where the agent learns to commit to the correct placement combinations for better task success. 
\begin{table}[t]
    \footnotesize
    \setlength\tabcolsep{1pt}
\resizebox{\columnwidth}{!}{
\begin{tabular}{@{}cllll@{}}
\toprule
\textbf{Scene Pairs} & \multicolumn{2}{c}{\textbf{Scene 1 \& 2}} & \multicolumn{2}{c}{\textbf{Scene 3 \& 4}} \\ \midrule
\textbf{Task Set} & \textit{\textbf{train(Scene2)}} & \textit{\textbf{test(Scene1)}} & \textit{\textbf{train(Scene4)}} & \textit{\textbf{test(Scene3)}} \\ \midrule
\textbf{LLM-Planner} & -8.4 ± 3.8 & -3.6 ± 4.8 & -29.6 ± 5.2 & -4.0 ± 3.3 \\ \midrule
\textbf{LLM-Personalize(\il{})} & -3.3 ± 3.0 & 13.2 ± 6.5 & 10.9 ± 3.3 & 34.3 ± 5.4 \\ \midrule
\textbf{LLM-Personalize(\si1)} & \textbf{19.4 ± 2.9} & 13.0 ± 4.6 & 24.2 ± 3.2 & 31.2 ± 4.9 \\ \midrule
\textbf{LLM-Personalize(\si2)} & 18.5 ± 2.6 & \textbf{17.0 ± 4.5} & \textbf{29.1 ± 2.9} & \textbf{42.4 ± 3.8} \\ \bottomrule
\end{tabular}
}
\caption{Ablation Study: Cross-domain generalisation success rate. (\textbf{Boldface}: best across the task set, \textbf{\si1/2}: self-training iteration 1 or 2, \textbf{\il{}}: imitation learning)}
\label{table:cross_domain}
\vspace{-5mm}
\end{table}

\section{CONCLUSIONS}
We proposed LLM-Personalize, a household robotics agent framework with an LLM-based architecture capable of performing long-horizon planning in multi-room, partially observable household scenarios, and an optimisation pipeline designed to personalize the LLM planner according to user preferences. This novel approach effectively addresses the critical gap in the personalization of LLM planners for household robotics. Our model achieves superior alignment with user preferences, outperforming existing work in the challenging Housekeep rearrangement tasks. Looking forward, the versatility and scalability of the agent design and optimization pipeline makes LLM-Personalize a promising solution for a broad range of household robotics applications.

\addtolength{\textheight}{-12cm}   




\bibliographystyle{IEEEtran}
\bibliography{reference}

\begin{thebibliography}{25}
\providecommand{\natexlab}[1]{#1}

\bibitem[{Ahn et~al.(2022)Ahn, Brohan, Brown, Chebotar, Cortes, David, Finn, Fu, Gopalakrishnan, Hausman et~al.}]{ahn2022can}
Michael Ahn, Anthony Brohan, Noah Brown, Yevgen Chebotar, Omar Cortes, Byron David, Chelsea Finn, Chuyuan Fu, Keerthana Gopalakrishnan, Karol Hausman, et~al. 2022.
\newblock Do as i can, not as i say: Grounding language in robotic affordances.
\newblock \emph{arXiv preprint arXiv:2204.01691}.

\bibitem[{Aky{\"u}rek et~al.(2023)Aky{\"u}rek, Aky{\"u}rek, Madaan, Kalyan, Clark, Wijaya, and Tandon}]{akyurek2023rl4f}
Afra~Feyza Aky{\"u}rek, Ekin Aky{\"u}rek, Aman Madaan, Ashwin Kalyan, Peter Clark, Derry Wijaya, and Niket Tandon. 2023.
\newblock Rl4f: Generating natural language feedback with reinforcement learning for repairing model outputs.
\newblock \emph{arXiv preprint arXiv:2305.08844}.

\bibitem[{Brown et~al.(2020)Brown, Mann, Ryder, Subbiah, Kaplan, Dhariwal, Neelakantan, Shyam, Sastry, Askell et~al.}]{brown2020language}
Tom Brown, Benjamin Mann, Nick Ryder, Melanie Subbiah, Jared~D Kaplan, Prafulla Dhariwal, Arvind Neelakantan, Pranav Shyam, Girish Sastry, Amanda Askell, et~al. 2020.
\newblock Language models are few-shot learners.
\newblock \emph{Advances in neural information processing systems}, 33:1877--1901.

\bibitem[{Chowdhery et~al.(2023)Chowdhery, Narang, Devlin, Bosma, Mishra, Roberts, Barham, Chung, Sutton, Gehrmann et~al.}]{chowdhery2023palm}
Aakanksha Chowdhery, Sharan Narang, Jacob Devlin, Maarten Bosma, Gaurav Mishra, Adam Roberts, Paul Barham, Hyung~Won Chung, Charles Sutton, Sebastian Gehrmann, et~al. 2023.
\newblock Palm: Scaling language modeling with pathways.
\newblock \emph{Journal of Machine Learning Research}, 24(240):1--113.

\bibitem[{Dong et~al.(2023)Dong, Xiong, Goyal, Pan, Diao, Zhang, Shum, and Zhang}]{dong2023raft}
Hanze Dong, Wei Xiong, Deepanshu Goyal, Rui Pan, Shizhe Diao, Jipeng Zhang, Kashun Shum, and Tong Zhang. 2023.
\newblock Raft: Reward ranked finetuning for generative foundation model alignment.
\newblock \emph{arXiv preprint arXiv:2304.06767}.

\bibitem[{Glaese et~al.(2022)Glaese, McAleese, Trebacz, Aslanides, Firoiu, Ewalds, Rauh, Weidinger, Chadwick, Thacker et~al.}]{glaese2022improving}
Amelia Glaese, Nat McAleese, Maja Trebacz, John Aslanides, Vlad Firoiu, Timo Ewalds, Maribeth Rauh, Laura Weidinger, Martin Chadwick, Phoebe Thacker, et~al. 2022.
\newblock Improving alignment of dialogue agents via targeted human judgements.
\newblock \emph{arXiv preprint arXiv:2209.14375}.

\bibitem[{Gulcehre et~al.(2023)Gulcehre, Paine, Srinivasan, Konyushkova, Weerts, Sharma, Siddhant, Ahern, Wang, Gu et~al.}]{gulcehre2023reinforced}
Caglar Gulcehre, Tom~Le Paine, Srivatsan Srinivasan, Ksenia Konyushkova, Lotte Weerts, Abhishek Sharma, Aditya Siddhant, Alex Ahern, Miaosen Wang, Chenjie Gu, et~al. 2023.
\newblock Reinforced self-training (rest) for language modeling.
\newblock \emph{arXiv preprint arXiv:2308.08998}.

\bibitem[{Huang et~al.(2022{\natexlab{a}})Huang, Abbeel, Pathak, and Mordatch}]{huang2022language}
Wenlong Huang, Pieter Abbeel, Deepak Pathak, and Igor Mordatch. 2022{\natexlab{a}}.
\newblock Language models as zero-shot planners: Extracting actionable knowledge for embodied agents.
\newblock In \emph{International Conference on Machine Learning}, pages 9118--9147. PMLR.

\bibitem[{Huang et~al.(2023)Huang, Xia, Shah, Driess, Zeng, Lu, Florence, Mordatch, Levine, Hausman et~al.}]{huang2023grounded}
Wenlong Huang, Fei Xia, Dhruv Shah, Danny Driess, Andy Zeng, Yao Lu, Pete Florence, Igor Mordatch, Sergey Levine, Karol Hausman, et~al. 2023.
\newblock Grounded decoding: Guiding text generation with grounded models for robot control.
\newblock \emph{arXiv preprint arXiv:2303.00855}.

\bibitem[{Huang et~al.(2022{\natexlab{b}})Huang, Xia, Xiao, Chan, Liang, Florence, Zeng, Tompson, Mordatch, Chebotar et~al.}]{huang2022inner}
Wenlong Huang, Fei Xia, Ted Xiao, Harris Chan, Jacky Liang, Pete Florence, Andy Zeng, Jonathan Tompson, Igor Mordatch, Yevgen Chebotar, et~al. 2022{\natexlab{b}}.
\newblock Inner monologue: Embodied reasoning through planning with language models.
\newblock \emph{arXiv preprint arXiv:2207.05608}.

\bibitem[{Kant et~al.(2022)Kant, Ramachandran, Yenamandra, Gilitschenski, Batra, Szot, and Agrawal}]{kant2022housekeep}
Yash Kant, Arun Ramachandran, Sriram Yenamandra, Igor Gilitschenski, Dhruv Batra, Andrew Szot, and Harsh Agrawal. 2022.
\newblock Housekeep: Tidying virtual households using commonsense reasoning.
\newblock In \emph{European Conference on Computer Vision}, pages 355--373. Springer.

\bibitem[{Liang et~al.(2023)Liang, Huang, Xia, Xu, Hausman, Ichter, Florence, and Zeng}]{liang2023code}
Jacky Liang, Wenlong Huang, Fei Xia, Peng Xu, Karol Hausman, Brian Ichter, Pete Florence, and Andy Zeng. 2023.
\newblock Code as policies: Language model programs for embodied control.
\newblock In \emph{2023 IEEE International Conference on Robotics and Automation (ICRA)}, pages 9493--9500. IEEE.

\bibitem[{Liu et~al.(2023)Liu, Sferrazza, and Abbeel}]{liu2023chain}
Hao Liu, Carmelo Sferrazza, and Pieter Abbeel. 2023.
\newblock Chain of hindsight aligns language models with feedback.
\newblock \emph{arXiv preprint arXiv:2302.02676}, 3.

\bibitem[{Mai et~al.(2023)Mai, Chen, Li, Qian, Elhoseiny, and Ghanem}]{mai2023llm}
Jinjie Mai, Jun Chen, Bing Li, Guocheng Qian, Mohamed Elhoseiny, and Bernard Ghanem. 2023.
\newblock Llm as a robotic brain: Unifying egocentric memory and control.
\newblock \emph{arXiv preprint arXiv:2304.09349}.

\bibitem[{Ouyang et~al.(2022)Ouyang, Wu, Jiang, Almeida, Wainwright, Mishkin, Zhang, Agarwal, Slama, Ray et~al.}]{ouyang2022training}
Long Ouyang, Jeffrey Wu, Xu~Jiang, Diogo Almeida, Carroll Wainwright, Pamela Mishkin, Chong Zhang, Sandhini Agarwal, Katarina Slama, Alex Ray, et~al. 2022.
\newblock Training language models to follow instructions with human feedback.
\newblock \emph{Advances in Neural Information Processing Systems}, 35:27730--27744.

\bibitem[{Pronobis and Jensfelt(2011)}]{pronobis2011hierarchical}
Andrzej Pronobis and Patric Jensfelt. 2011.
\newblock Hierarchical multi-modal place categorization.
\newblock In \emph{ECMR}, pages 159--164.

\bibitem[{Rafailov et~al.(2023)Rafailov, Sharma, Mitchell, Ermon, Manning, and Finn}]{rafailov2023direct}
Rafael Rafailov, Archit Sharma, Eric Mitchell, Stefano Ermon, Christopher~D Manning, and Chelsea Finn. 2023.
\newblock Direct preference optimization: Your language model is secretly a reward model.
\newblock \emph{arXiv preprint arXiv:2305.18290}.

\bibitem[{Rafailov et~al.(2024)Rafailov, Sharma, Mitchell, Manning, Ermon, and Finn}]{rafailov2024direct}
Rafael Rafailov, Archit Sharma, Eric Mitchell, Christopher~D Manning, Stefano Ermon, and Chelsea Finn. 2024.
\newblock Direct preference optimization: Your language model is secretly a reward model.
\newblock \emph{Advances in Neural Information Processing Systems}, 36.

\bibitem[{Rana et~al.(2023)Rana, Haviland, Garg, Abou-Chakra, Reid, and Suenderhauf}]{rana2023sayplan}
Krishan Rana, Jesse Haviland, Sourav Garg, Jad Abou-Chakra, Ian Reid, and Niko Suenderhauf. 2023.
\newblock Sayplan: Grounding large language models using 3d scene graphs for scalable task planning.
\newblock \emph{arXiv preprint arXiv:2307.06135}.

\bibitem[{Scheurer et~al.(2023)Scheurer, Campos, Korbak, Chan, Chen, Cho, and Perez}]{scheurer2023training}
J{\'e}r{\'e}my Scheurer, Jon~Ander Campos, Tomasz Korbak, Jun~Shern Chan, Angelica Chen, Kyunghyun Cho, and Ethan Perez. 2023.
\newblock Training language models with language feedback at scale.
\newblock \emph{arXiv preprint arXiv:2303.16755}.

\bibitem[{Song et~al.(2023)Song, Wu, Washington, Sadler, Chao, and Su}]{song2023llm}
Chan~Hee Song, Jiaman Wu, Clayton Washington, Brian~M Sadler, Wei-Lun Chao, and Yu~Su. 2023.
\newblock Llm-planner: Few-shot grounded planning for embodied agents with large language models.
\newblock In \emph{Proceedings of the IEEE/CVF International Conference on Computer Vision}, pages 2998--3009.

\bibitem[{Touvron et~al.(2023)Touvron, Martin, Stone, Albert, Almahairi, Babaei, Bashlykov, Batra, Bhargava, Bhosale et~al.}]{touvron2023llama}
Hugo Touvron, Louis Martin, Kevin Stone, Peter Albert, Amjad Almahairi, Yasmine Babaei, Nikolay Bashlykov, Soumya Batra, Prajjwal Bhargava, Shruti Bhosale, et~al. 2023.
\newblock Llama 2: Open foundation and fine-tuned chat models.
\newblock \emph{arXiv preprint arXiv:2307.09288}.

\bibitem[{Vaswani et~al.(2017)Vaswani, Shazeer, Parmar, Uszkoreit, Jones, Gomez, Kaiser, and Polosukhin}]{vaswani2017attention}
Ashish Vaswani, Noam Shazeer, Niki Parmar, Jakob Uszkoreit, Llion Jones, Aidan~N Gomez, {\L}ukasz Kaiser, and Illia Polosukhin. 2017.
\newblock Attention is all you need.
\newblock \emph{Advances in neural information processing systems}, 30.

\bibitem[{Wu et~al.(2023)Wu, Antonova, Kan, Lepert, Zeng, Song, Bohg, Rusinkiewicz, and Funkhouser}]{wu2023tidybot}
Jimmy Wu, Rika Antonova, Adam Kan, Marion Lepert, Andy Zeng, Shuran Song, Jeannette Bohg, Szymon Rusinkiewicz, and Thomas Funkhouser. 2023.
\newblock Tidybot: Personalized robot assistance with large language models.
\newblock \emph{arXiv preprint arXiv:2305.05658}.

\bibitem[{Xu et~al.(2022)Xu, He, He, and McAuley}]{xu2022leashing}
Canwen Xu, Zexue He, Zhankui He, and Julian McAuley. 2022.
\newblock Leashing the inner demons: Self-detoxification for language models.
\newblock In \emph{Proceedings of the AAAI Conference on Artificial Intelligence}, volume~36, pages 11530--11537.

\end{thebibliography}

\end{document}